\documentclass[lettersize,journal]{IEEEtran}
\usepackage{amsmath,amsfonts}
\usepackage{algorithmic}
\usepackage{algorithm}
\usepackage{array}
\usepackage[caption=false,font=footnotesize,margin=10pt]{subfig}
\usepackage{textcomp}
\usepackage{stfloats}
\usepackage{url}
\usepackage{verbatim}
\usepackage{graphicx}
\usepackage{cite}
\usepackage{algorithm,algorithmic}
\usepackage{xcolor}
\usepackage{float}
\usepackage{multirow}
\usepackage{makecell}
\usepackage{multirow}
\usepackage{environ}
\usepackage{pbalance}
\usepackage{orcidlink}

\usepackage{tikz}
\usetikzlibrary{positioning, calc}
\usetikzlibrary{arrows.meta}

\newcommand{\mb}[1]{\mathbf{#1}}
\newcommand{\real}[1]{\mathbb{R}^{#1}}
\newcommand{\bm}[1]{\begin{bmatrix}
		#1
\end{bmatrix}}
\newcommand{\mcal}[1]{\mathcal{#1}}
\newcommand{\ldef}{\stackrel{\Delta}{=}}

\newcommand{\expect}[2]{\mathbb{E}_{#1} \left[ #2 \right]}
\newcommand{\bt}[1]{\left[ \texttt{#1}\right]}

\usepackage[most]{tcolorbox}
\tcbset{highlight math style={enhanced jigsaw,size=tight,colback=light-gray,boxrule=0pt,round corners}}
\newcommand{\bluebg}[1]{%
	\colorbox{blue!20}{#1}
}
\newcommand{\redbg}[1]{%
	\colorbox{red!20}{#1}
}
\newcommand{\greenbg}[1]{%
	\colorbox{green!20}{#1}
}
\newcommand{\magbg}[1]{%
	\colorbox{magenta!30}{#1}
}
\newcommand{\orbg}[1]{%
	\colorbox{orange!30}{#1}
}

\usetikzlibrary{positioning, calc}
\usetikzlibrary{arrows.meta}
\usetikzlibrary{positioning,shapes,shadows,arrows}

\tikzstyle{abstract}=[rectangle, draw=black, rounded corners, fill=white, drop shadow,
text centered, anchor=north, text width=8cm]
\tikzstyle{abstract2}=[rectangle, draw=black, rounded corners, fill=white, drop shadow,
text centered, anchor=north, text width=8cm]
\tikzstyle{myarrow}=[->, >=open triangle 90, thick]

\makeatletter
\newsavebox{\measure@tikzpicture}
\NewEnviron{scaletikzpicturetowidth}[1]{%
	\def\tikz@width{#1}%
	\begin{lrbox}{\measure@tikzpicture}%
		\BODY
	\end{lrbox}%
	\pgfmathparse{#1/\wd\measure@tikzpicture}%
	\BODY
}
\makeatother

\begin{document}
	
\title{Federated Short-Term Load Forecasting with \\Personalization Layers for Heterogeneous Clients}
	
\author{
Shourya Bose\orcidlink{0000-0002-2081-9545},~\IEEEmembership{Graduate Student Member,~IEEE}, and
Kibaek~Kim\orcidlink{0000-0002-5820-6533},~\IEEEmembership{Senior Member,~IEEE}
\thanks{S. Bose is with the Department of Electrical and Computer Engineering, University of California, Santa Cruz, CA, USA (e-mail: shbose@ucsc.edu).}
\thanks{K. Kim is with the Mathematics and Computer Science Division, Argonne National Laboratory, Lemont, IL, USA (e-mail: kimk@anl.gov).}
\thanks{This material is based upon work supported by the U.S. Department of Energy, Office of Science, under contract number DE-AC02-06CH11357. We gratefully acknowledge the computing resources provided on Bebop and Swing, a high-performance computing cluster operated by the Laboratory Computing Resource Center at Argonne National Laboratory.}
}
	
	
	\maketitle
	
	\begin{abstract}
		The advent of smart meters has enabled pervasive collection of energy consumption data for training short-term load forecasting (STLF) models. In response to privacy concerns, federated learning (FL) has been proposed as a privacy-preserving approach for training, but the quality of trained models degrades as client data becomes heterogeneous. In this paper we alleviate this drawback using \emph{personalization layers}, wherein certain layers of an STLF model in an FL framework are trained exclusively on the clients' own data. To that end, we propose a personalized FL algorithm---PL-FL---enabling FL to handle personalization layers. The PL-FL algorithm is implemented by using the Argonne Privacy-Preserving Federated Learning  package. We test the forecast performance of models trained on the NREL ComStock dataset, which contains heterogeneous energy consumption data of multiple commercial buildings. Superior performance of models trained with PL-FL demonstrates that personalization layers enable classical FL algorithms to handle clients with heterogeneous data.
	\end{abstract}
	
	
	\section{Introduction}
	
	Short-term load forecasting (STLF) is  important for the operation of electric grids~\cite{ME-JAKS-RB-BDM:2007}. It refers to forecasting the consumption of electrical energy by an entity over a duration ranging from a few minutes to a day. The input for generating forecasts includes past energy consumption and other data such as date-time indices or weather metrics that are predictive of future energy consumption. Accurate forecasting enables electrical utilities to plan the operation of generation units in an economical fashion. Coupled with high-quality forecasts of renewable energy generation such as solar and wind, STLF plays an important role in the integration of distributed renewable generation into the grid. One can  show that the accuracy of STLF models has a direct and significant effect on reliability of electricity supply~\cite{BW-YL-JW:2013}.
	
	Most STLF models process historical consumption data to learn relevant patterns, which are then used to generate forecasts. Two important characteristics of such models are temporal granularity (referring to the time interval between successive forecast points) and aggregation level (referring to the number of entities whose consumption is summed at each forecast point). Traditionally, STLF models have had low temporal granularity of approximately 1 hour or greater~\cite{GG-FDG:1987} and large aggregation levels at the scale of entire neighborhoods~\cite{HLW-JEDNG:1983}. Over the past two decades, many utility companies have adopted the paradigm of smart grid~\cite{XY-CC-TD-GM:2011}, which involves residences and commercial buildings equipped with advanced metering infrastructure (AMI) such as smart meters~\cite{YK:2016}. AMI allows for monitoring energy consumption on a timescale ranging over 1 minute to 1 hour and at aggregation levels corresponding to individual households or buildings. Considering the availability of such data, there has been significant research into the use of deep learning (DL)-based models such as long short-term memory (LSTM), which can be leveraged to provide accurate forecasts~\cite{WK-etal:2019}. DL-based models are the current state of the art in STLF~\cite{TH-JX-JB:2019}, although state-space-based autoregressive models such as SARIMAX~\cite{AT-ILA:2017} remain a popular alternative.
	
	However, the availability of such fine-grained data has raised significant concerns regarding privacy of individuals or corporations whose residences' or buildings' electricity consumption is monitored through AMI. Studies by Molina--Markham et al.~\cite{AMM-PS-KF-EC-DI:2010} and Beckel et al.~\cite{CB-LS-TS-SS:2014} show that residential smart meter data can allow inference of occupants' personal data such social class, employment status, and marital status using standard statistical or DL techniques. Furthermore, it increases the risk of industrial espionage. Considering these pitfalls, several governments have implemented legislation that seeks to preserve consumer privacy in face of pervasive ingestion of smart meter data by utilities and third parties. For example, General Data Protection Regulation  and its various implementations in Europe and AB 1274 in California either recommend or mandate the use of privacy-preserving practices on data collected from smart meters~\cite{DL-DJH:2021}. This covers usage of smart meter data for creating of STLF models. 
	
	\emph{Federated learning} (FL)~\cite{KB-etal:2017} is an emerging framework to address this challenge, wherein distributed training of models using local data on edge devices preserves privacy through data compartmentalization. However, many popular FL algorithms suffer from performance degradation when edge devices (henceforth called \emph{clients}) contain heterogeneous data~\cite{TL-AKS-AT-VS:2020}. This is especially relevant when using FL to train STLF models (wherein clients represent individual smart meters), since the electricity consumptions of different households and buildings follow significantly different patterns and may not be strongly correlated with each other. We address this issue through the recently introduced concept of \emph{personalization layers}~\cite{MGA-VA-AKS:2019,LC-HH-AM-SS:2021,XM-JZ-SG-WX:2022}. Personalization layers split the layers of the DL model into two disjoint subsets: shared layers and  personalized layers. Personalized layer weights remain local to the client, while shared layer weights are sent to the aggregating server (henceforth simply called \emph{server}) in order to exploit the benefits of FL. Recent encouraging results in image classification tasks~\cite{LC-HH-AM-SS:2021} serve as a motivation to try these  techniques for STLF.
	
\subsubsection*{Literature Review of STLF} 
Significant recent progress has been made in the application of DL for STLF. Several DL models such as multilayer perceptron~\cite{Chen-etal:1992}, convolutional neural networks~\cite{MV-etal:2018}, residual neural networks~\cite{KC-etal:2019}, gated recurrent unit (with~\cite{SJ-etal:2021} and without~\cite{MX-etal:2021} attention mechanism~\cite{attention-2017}), and LSTM (with~\cite{JL-etal:2022} and without~\cite{WK-etal:2019} attention mechanism) have shown impressive performance in STLF on various energy consumption datasets.
 
Motivated by privacy concerns arising from the need of large datasets to train the aforementioned models, recent literature has attempted to capture various use cases of FL in STLF. Ta\"ik and Cherkaoui~\cite{AT-SC:2020} present an FL scheme based on the federated averaging (\emph{FedAvg}) algorithm for training an LSTM forecasting model. A similar setting is chosen by Fekri et al.~\cite{MNF-KG-SM:2022}, except that the authors consider \emph{FedAvg} as well as \emph{FedSGD} algorithms. Fern\'{a}ndez et al.~\cite{JDF-etal:2022} provide a comprehensive experimental review of various FL techniques for residential STLF, including  the aforementioned algorithms, various STLF model architectures, and additional privacy-preserving techniques such as differential privacy~\cite{CD-AR:2014}. Chen et al.~\cite{ZC-JL-XC-XL:2023} consider the problem of generating synthetic data for training STLF models and address it by designing a federated generative adversarial network  with a centralized generator and federated discriminators. Husnoo et al.~\cite{MAH-etal:2023} devise a federated STLF scheme with quantized communication and differential privacy that is robust to adversarial clients orchestrating \emph{Byzantine attacks} to cause data leakage of other clients' data. 
	
However, the heterogeneity of the clients' local energy consumption data poses a challenge to the aforementioned FL methods. Several recent works  propose solutions to improve FL performance under heterogeneity. Su et al.~\cite{ZS-etal:2022} consider a setting where clients with heterogeneous data are incentivized by the server to share their data. In this setting, the authors propose a reinforcement learning  framework that allows the server to collect data from the clients in a way that optimizes the resulting accuracy. Wang et al.~\cite{YW-etal:2023} train an STLF model on pooled data of all clients, followed by local training epochs on each client's data to generated personalized models. A similar strategy is employed by Grabner et al.~\cite{MG-YW-QW-BB-VS:2023}, who further use ensembles of similar clients' models to improve forecasting accuracy.  Qin et al.~\cite{DQ-etal:2023} consider STLF in a metalearning framework wherein clients locally determine an optimal model architecture, followed by clustering of clients with similar models, and each cluster jointly trains an STLF model. Even outside the context of FL, Weicong et al.~\cite{WK-etal:2019} recognize the challenge of heterogeneity in STLF datasets and use a clustering algorithm to group similar users' data before training different STLF models for each group.
	
Current literature on personalizing FL for STLF models is limited to a priori clustering and fine-tuning of FL models through additional client-level training epochs~\cite{MG-YW-QW-BB-VS:2023,YW-etal:2023}. Contrary to this approach, we demonstrate the use of personalization layers for combating data heterogeneity while training an STLF model in the federated setting. We restrict our attention to one-step-ahead forecasting, since this allows us to interpret the results of our experiments with greater clarity. We use the LSTM-based model proposed by Weicong et al.~\cite{WK-etal:2019} for this task, since this model demonstrated better performance on the heterogeneous datasets used for the experiments in this paper comopared with other models including attention-based versions of LSTM.
 
\subsubsection*{Contribution}
	
We first present three variants of the FL approach to train STLF models as an algorithm. We then modify the FL algorithm to personalize the model locally at each client by excluding certain layers of the model from federation. We call the resulting approach PL-FL (denoting \emph{personalization layers FL}) and hypothesize that this can resolve the performance degradation due to clients' data heterogeneity. This hypothesis is tested on three different datasets from the NREL ComStock data repository~\cite{ComStock}, corresponding to three U.S. states with 42 commercial buildings in each. We explore the heterogeneity in the dataset, followed by carrying out two experiments to establish the best algorithmic and personalization configuration for the LSTM-based STLF model. We use the Argonne Privacy-Preserving Federated Learning (APPFL)~\cite{APPFL} package to run both a standard FL (i.e., without the personalization) and PL-FL on a cluster with multiple GPUs and multicore CPUs. The obtained results demonstrate the effectiveness of PL-FL in addressing client data heterogeneity.
	
\subsubsection*{Organization}

Section~\ref{sec:sysmodel} introduces the architecture of the LSTM-based STLF model, followed by the presentation and discussion of both FL and the personalized variation PL-FL. Section~\ref{sec:exploratory} highlights the rationale for including various features in the STLF model by exploring the correlation of energy consumption with other features. Furthermore, this section highlights the heterogeneity in the dataset through box plots. Section~\ref{sec:experiments} presents two experiments: the first  explores the ideal server algorithm for federated STLF, and the second compares PL-FL with three personalization configurations with FL and non-federated training.  Section~\ref{sec:conclusion}  summarizes our conclusions and briefly mentions ufture work.
	
\subsubsection*{Notation} 

$\real{}$ denotes the real numbers. For a positive integer $a$, $[a]$ and $[a]_0$ denote the sets $\{ 1,\cdots,a\}$ and $\{0,\cdots,a\}$ respectively. For a finite set $\mcal{D}$, $|\mcal{D}|$ denotes its cardinality. Vectors and matrices are denoted in boldface. For vectors $\mb{a},\mb{b}\in\real{n}$, $\mb{a}\odot\mb{b}$ denotes their Hadamard (elementwise) product, while $\mb{a}^{\odot c}$ denotes elementwise exponentiation of $\mb{a}$ to some power $c\in\real{}$. $\expect{\mb{x}\sim \mcal{P}}{f(\mb{x})}$ denotes the expected value of function $f (\, .\,)$ when its argument is sampled from the probability distribution $\mcal{P}$. For a continuous function $f:\real{n}\mapsto \real{}$, $\nabla_{\mb{x}} f(\mb{x'})$ denotes the gradient (or any subgradient, in case the gradient is not defined) of $f$ at point $\mb{x}'$. 
	
\section{System Model}
\label{sec:sysmodel}
	
In this section we describe the LSTM-based STLF model, followed by introduction of FL and PL-FL algorithms for training this model on training data of clients in a federated setting.
	
\subsection{Long Short-Term Memory }
	
We adopt the LSTM model presented in~\cite{WK-etal:2019} for one-step-ahead load forecasting.
The purpose of an LSTM model is to produce an output, given a sequence of inputs $\{ \mb{x}_0,\mb{x}_1,\cdots,\mb{x}_{T-1}\}$, wherein $\mb{x}_t\in\real{l}$ for time period $t\in[T-1]_0$ and $l>0$ denotes the number of features contained in the input. Fundamental to any LSTM model is the LSTM cell, which is iterated over sequential inputs and two running cell state variables. The LSTM cell ingests states denoted by $\mb{s}_{t-1},\mb{h}_{t-1}\in\real{m}$ and input element $\mb{x}_{t-1}$ and outputs the updated cell states $\mb{s}_{t}$ and $\mb{h}_{t}$. These states can then be fed back into the LSTM cell along with the next input element $\mb{x}_{t}$, and the process continues. Letting $\sigma (\, . \,)$ and $\phi(\, .\,)$ denote the elementwise sigmoid and tanh functions, respectively, the internal structure of the cell is described by the following equations:
\begin{subequations}
    \label{eq:lstm_equations}
    \begin{align}
        \mb{f}_{t} &= \sigma(\mb{W}_{fx}\mb{x}_{t-1} + \mb{W}_{fh}\mb{h}_{t-1} + \mb{b}_f)\\
        \mb{i}_{t} &= \sigma( \mb{W}_{ix} \mb{x}_{t-1} + \mb{W}_{ih}\mb{h}_{t-1} + \mb{b}_i)\\
        \mb{g}_{t} &= \phi(\mb{W}_{gx}\mb{x}_{t-1} + \mb{W}_{gh}\mb{h}_{t-1} + \mb{b}_g)\\
        \mb{o}_{t} &= \sigma(\mb{W}_{ox}\mb{x}_{t-1} + \mb{W}_{oh}\mb{h}_{t-1} + \mb{b}_o)\\
        \mb{s}_{t} &= \mb{g}_{t} \odot \mb{i}_{t} + \mb{s}_{t-1} \odot \mb{f}_{t}\\
        \mb{h}_t &= \phi(\mb{s}_t) \odot \mb{o}_{t}.
    \end{align}
\end{subequations}

In equations~\eqref{eq:lstm_equations}, the learnable parameters (i.e., the parameters that are updated as a function of data during training) are the weights $\mb{W}_{(\, . \,)}$ and biases $\mb{b}_{(\, . \,)}$. Initializing the cell states as zero,~\eqref{eq:lstm_equations} may be represented as function $L$ with the input sequence $\{\mb{x}_0,\cdots,\mb{x}_{T-1}\}$ and the output sequence $\{ \mb{h}_1,\cdots,\mb{h}_T\}$:
\begin{gather*}
    \mb{s}_{t},\mb{h}_{t} = L\left(\mb{s}_{t-1},\mb{h}_{t-1},\mb{x}_{t-1} | \mb{W}_{(\, . \,)}, \mb{b}_{(\, . \,)}\right),\forall t\in[T],\\
    \mb{h}_0 = \mb{0}_m,\mb{s}_0 = \mb{0}_m.
\end{gather*}

Multiple LSTM layers may be stacked on top of each other. For example, a two-stacked LSTM layer can be written as follows.

\begin{gather*}
    \mb{s}^{(1)}_{t},\mb{h}^{(1)}_{t} = L\left(\mb{s}^{(1)}_{t-1},\mb{h}^{(1)}_{t-1},\mb{x}_{t-1} | \mb{W}^{(1)}_{(\, . \,)}, \mb{b}^{(1)}_{(\, . \,)}\right),\forall t\in[T],\\
    \mb{s}^{(2)}_{t},\mb{h}^{(2)}_{t} = L\left(\mb{s}^{(2)}_{t-1},\mb{h}^{(2)}_{t-1},\mb{h}^{(1)}_{t} | \mb{W}^{(2)}_{(\, . \,)}, \mb{b}^{(2)}_{(\, . \,)}\right),\forall t\in[T],\\
    \mb{h}^{(1)}_0 = \mb{0}_m,\mb{s}^{(1)}_0 = \mb{0}_m, \mb{h}^{(2)}_0=\mb{0}_m, \mb{s}^{(2)}_0 = \mb{0}_m
\end{gather*}

\begin{figure}[!tb]
    \centering
    \includegraphics[width=\linewidth]{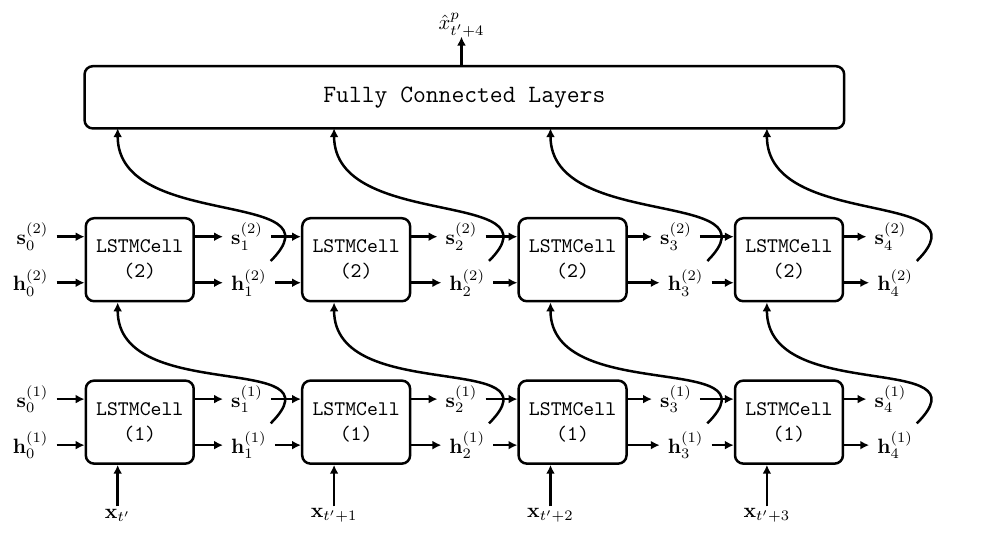}
    \caption{Schematic of LSTM model for STLF. In this example the lookback duration is $T=4$.}
    \label{fig:lstm1}
\end{figure}

We now specialize the LSTM architecture for one-step-ahead STLF according to~\cite{WK-etal:2019}. Each element of the input sequence may be written as $\mb{x}_t = \bm{ x^p_t & \left(\mb{x}^f_t\right)^\top}^\top$, where $x^p_t\in\real{}$ is the energy consumption on past timestep $t$ and $\mb{x}^f_t\in\real{l-1}$ denotes $l-1$ additional features such as date/time indices and weather data. The contents of these features are discussed in depth in Section~\ref{sec:exploratory}. We use a two-stacked LSTM model, and the $\mb{h}$-states of the top layer for all timesteps are concatenated and fed into a fully connected module. This module has three linear layers separated by parametric rectified linear unit (PReLU) activation functions. The output of the fully connected module is $\hat{x}_{T}^p$, representing the estimate of energy consumption on time step $T$. A schematic of the proposed architecture for $T=4$ and time offset $t'$ is presented in Figure~\ref{fig:lstm1}.

\subsection{Federated Learning for STLF}\label{ssec:fl}


We consider an electrical utility company (server) that aims to develop accurate load forecasting models for $M$ customers (clients) equipped with smart meters. Each client $m$ possesses a historical energy consumption dataset of the form $\mcal{D}_m \ldef\{\mb{X}_i,Y_i\}_{i=1}^{N_m}$, wherein 
\begin{align*}
    \mb{X}_i = \{ \mb{x}_{t'},\cdots,\mb{x}_{t'+T-1}\}, \quad Y_i = x^p_{t'+T}
\end{align*}
denote the input and expected output for the STLF model and $i$ is used to index the $N_m$ dataset elements of client $m$. The time index $t'$ can take any past value contained in the dataset. We assume that the dataset of client $m$ is sampled in an independent and identically distributed (i.i.d) fashion from a probability distribution $\mcal{P}_m$. $T$ is known as the \emph{lookback} duration and represents the number of previous data points used to generate the forecast. We use $f_{\pmb{\theta}}(\,.\,)$ to denote the LSTM-based STLF model introduced in the preceding subsection, wherein $\pmb{\theta}$ denotes the learnable weights $\mb{W}^{(1)}_{(\, . \,)},\mb{W}^{(2)}_{(\, . \,)},\mb{b}^{(1)}_{(\, . \,)},\mb{b}^{(2)}_{(\, . \,)}$ and the weights and biases of the fully connected layers. We use the mean-squared error  as the training loss metric, which is the average of squared error loss $l(x,y) \ldef (x-y)^2$ over multiple data points. The goal of training the model with FL is to solve the problem
\begin{align}
    \label{eq:fl-objective}
    \min\limits_{\pmb{\theta}} \left\{ \frac{1}{M} \sum_{m=1}^M \expect{(\mb{X},Y)\sim\mcal{P}_m}{l(f_{\pmb{\theta}}(\mb{X}),Y )}  \right\}.
\end{align}
However, the expectation in~\eqref{eq:fl-objective} is not computable in practice since $\mcal{P}_m$ is unknown, and therefore we approximate the expectation with a sample average over dataset $\mcal{D}_m$ as 
\begin{align}
    \label{eq:fl-objective2}
    \min\limits_{\pmb{\theta}} \left\{ \frac{1}{M} \sum_{m=1}^M \frac{1}{|\mcal{D}_m|} \sum_{(\mb{X},Y)\in\mcal{D}_m} l(f_{\pmb{\theta}}(\mb{X}),Y) \right\}.
\end{align}
Problem~\eqref{eq:fl-objective2} can be solved in the FL framework as described in Algorithm~\ref{alg:cent-fl}. It consists of the server maintaining a copy of model weights, which are distributed to all clients every server epoch. The clients perform a fixed number of gradient-based update epochs on the received weights with respect to minibatches sampled from their local dataset. The updated weights are communicated back to the server, which combines them to form the new server weights. The process continues for a fixed number of epochs.

In this paper, we use the Adam optimizer~\cite{DPK-JB:2014} for the client update epochs, which is a popular algorithm for unconstrained minimization. Compared to gradient descent which updates weights by adding to them the negatively scaled gradient of the loss function, Adam maintains two additional states denoted by $\hat{\mb{m}}$ and $\hat{\mb{v}}$. These states can be thought of as running low-pass filters over the first and second moments of the gradients respectively. Incorporating these states in the weight update smooths fluctuations in the gradients and provides per-weight learning rate adaptation, thereby stabilizing the clients' local optimization.

We experiment three FL server algorithms for the server updates: FedAve, FedAvgMomentum, and FedAdam. FedAvg can be thought of as simply averaging the weights received from all the clients with no smoothing whatsoever (line 21, Algorithm~\ref{alg:cent-fl}) and using it to update server weights. The popular FedSGD algorithm is closely related to FedAvg; indeed, in case of single epoch client updates, FedAvg and FedSGD are equivalent~\cite{DC-LW-LY-JZ-KC-QY:2022}. On the other hand, FedAvgMomentum maintains a smoothing state which considers the first moment of the clients' gradients (lines 22-23, Algorithm~\ref{alg:cent-fl}), while FedAdam mantains smoothing states for both first and second moments of the clients' gradients (lines 24-26, Algorithm~\ref{alg:cent-fl}). The state tracking the gradient's second moment in FedAdam provides per-weight adaptivity to server updates, similar to Adam. Note that every server-client interaction involves exchanging the entire weights contained in $\pmb{\theta}$, and therefore communication costs would increase rapidly with increasing size and complexity of STLF models~\cite{KB-etal:2019}.

\begin{algorithm}[!tb]
    \caption{FL with Adam optimizer at client and one of \redbg{FedAvg}, \greenbg{FedAvgMomentum}, or \bluebg{FedAdam} algorithms at server }
    \begin{algorithmic}[1]
        \renewcommand{\algorithmicrequire}{\textbf{Input:}}
        \renewcommand{\algorithmicensure}{\textbf{Output:}}
        \REQUIRE Datasets $\mcal{D}_m$ for clients $m\in[M]$, Server epochs $K$, Client epochs $\tilde{K}$, Server learning rate $\eta$, Client learning rate $\tilde{\eta}$, Client parameters $\tilde{\beta_1},\tilde{\beta}_2\in[0,1)$, Client adaptivity param. $\tilde{\epsilon}>0$, Client minibatch sampling rule and size\\
        \greenbg{Server parameter $\beta_1 \in [0,1)$}\\
        \bluebg{Server parameters $\beta_1,\beta_2\in[0,1)$,}\\
        \bluebg{Server adaptivity param. $\epsilon>0$}\\
        \ENSURE Trained centralized FL model $\pmb{\theta}^*$
        \STATE Initialize weights $\pmb{\theta}_0$
        \STATE \greenbg{Initialize FedAvgMomentum state $\mb{m}_0 = \mb{0}$}
        \STATE \bluebg{Initialize FedAdam states $\mb{m}_0=\mb{0}, \mb{v}_0=\epsilon^2\mb{1}$}
        \FOR {server epochs $k = 1$ to $K$}
        \STATE Server sends $\pmb{\theta}_{k-1}$ to all clients 
        \FOR {clients $m=1$ to $M$}
        \STATE Sample minibatch $\mcal{M}_m \subset \mcal{D}_m$
        \STATE Initialize local weights $\tilde{\pmb{\theta}}_{0,k}^m = \pmb{\theta}_{k-1}$
        \STATE Initialize Adam states $\mb{m}_{0,k}=\mb{0}, \mb{v}_{0,k}=\mb{0}$
        \\ \textit{Client Updates} :
        \FOR {client epochs $k'=1$ to $K'$}\label{line:CentClientStart}
        \STATE $\tilde{\mb{g}}_{k',k}^m = \nabla_{\pmb{\theta}} \frac{1}{|\mcal{M}_m|}\sum_{(\mb{X},Y)\in\mcal{M}_m}l(f_{\tilde{\pmb{\theta}}_{k'-1,k}}(\mb{X},Y))$
        \STATE $\mb{m}_{k',k}^m = \tilde{\beta}_1\mb{m}_{k'-1,k}^m + (1-\tilde{\beta}_1) \tilde{\mb{g}}_{k',k}^m$
        \STATE $\mb{v}_{k',k}^m = \tilde{\beta}_2\mb{v}_{k'-1,k}^m + (1-\tilde{\beta}_2)(\tilde{\mb{g}}_{k',k}^m)^{\odot 2} $
        \STATE $\mb{\hat{m}}_{k',k}^m = \mb{m}_{k',k}^m(1-\tilde{\beta}_1^{k'})^{-1}$
        \STATE $\mb{\hat{v}}_{k',k}^m = \mb{v}_{k',k}^m(1-\tilde{\beta}_2^{k'})^{-1}$
        \STATE $\tilde{\pmb{\theta}}_{k',k}^m = \tilde{\pmb{\theta}}_{k'-1,k}^m - \tilde{\eta}\mb{\hat{m}}_{k',k}^m\odot ((\mb{\hat{v}}_{k',k}^m)^{\odot\frac{1}{2}}+\tilde{\epsilon}\mb{1})^{\odot -1}$
        \ENDFOR \label{line:CentClientEnd}
        \STATE Client sends $\tilde{\mb{g}}^m_k=\pmb{\theta}_{k-1} - \tilde{\pmb{\theta}}^m_{K',k}$ to server
        \ENDFOR
        \\ \textit{Server Updates} :
        \STATE $\Delta_k =\sum_{m=1}^M \left( \frac{|\mcal{M}_m|}{\sum_{m'=1}^M |\mcal{M}_{m'}|} \right) \tilde{\mb{g}}^m_k $\label{line:CentServerStart}
        \STATE \redbg{$\pmb{\theta}_k =  \pmb{\theta}_{k-1} - \eta\Delta_k$}
        \STATE \greenbg{$\mb{m}_k = \beta_1\mb{m}_{k-1} + (1-\beta_1)\Delta_k$}
        \STATE \greenbg{$\pmb{\theta}_k = \pmb{\theta}_{k-1} - \eta\mb{m}_k$}
        \STATE \bluebg{$\mb{m}_k = \beta_1\mb{m}_{k-1}+(1-\beta_1)\Delta_k$ }
        \STATE \bluebg{$\mb{v}_k = \beta_2\mb{v}_{k-1} + (1-\beta_2)\Delta_k^{\odot 2}$}
        \STATE \bluebg{$\pmb{\theta}_k = \pmb{\theta}_{k-1}-\eta \mb{m}_k \odot(\mb{v}_k^{\odot \frac{1}{2}}+\epsilon\mb{1})^{\odot -1}$}\label{line:CentServerEnd}
        \ENDFOR
        \RETURN $\pmb{\theta}^*=\pmb{\theta}_K$ 
    \end{algorithmic} 
    \label{alg:cent-fl}
\end{algorithm}

\subsection{FL with Personalization Layers for STLF}

We now modify Algorithm~\ref{alg:cent-fl} to incorporate personalized layers for each client. Suppose the weights $\pmb{\theta}$ can be written as $\pmb{\theta} = [\pmb{\phi},\pmb{\psi}]$, wherein $\pmb{\phi}$ and $\pmb{\psi}$ correspond to the weights of the federated and personalized layers, respectively. In this setting, each client $m$ maintains its own copy of $\pmb{\psi}^m$ which is not communicated with the server, while the server maintains a global copy of $\pmb{\phi}$. Problem~\eqref{eq:fl-objective2} can now be written as follows.
\begin{align}
    \label{eq:fl-objective3}
    \min\limits_{\pmb{\phi},\{\pmb{\psi}^m\}_{m=1}^M} \left\{ \frac{1}{M} \sum_{m=1}^M \frac{1}{|\mcal{D}_m|} \sum_{(\mb{X},Y)\in\mcal{D}_m} l(f_{\pmb{\phi},\pmb{\psi}^m}(\mb{X}),Y)\right\}
\end{align}
We propose Algorithm~\ref{alg:pers-fl} to solve problem~\eqref{eq:fl-objective3}, which we refer to as PL-FL. One of the major benefits of PL-FL over FL is that personalizing certain layers in the STLF model allows it to better learn data from heterogenous clients, since the personalized layer weights change only as a function of the clients' local data. This is shown to be empirically true for STLF in Section~\ref{sec:experiments}. Another important advantage is the reduced communication cost. In FL, each client has to exchange the full parameter $\pmb{\theta}$ on each server epoch, while in PL-FL, the only communications needed are that of the shared layer weights $\pmb{\phi}$, which is a subset of $\pmb{\theta}$. This is an important step in adapting FL for use in smart meters, since most smart meters communicate with the server over cellular wireless networks and may therefore be bandwidth-constrained.

\begin{algorithm}[!tb]
    \caption{PL-FL with Adam optimizer at client and one of \redbg{FedAvg}, \greenbg{FedAvgMomentum}, or \bluebg{FedAdam} algorithms at server }
    \begin{algorithmic}[1]
        \renewcommand{\algorithmicrequire}{\textbf{Input:}}
        \renewcommand{\algorithmicensure}{\textbf{Output:}}
        \REQUIRE Datasets $\mcal{D}_m$ for clients $m\in[M]$, Server epochs $K$, Client epochs $\tilde{K}$, Server learning rate $\eta$, Client learning rate $\tilde{\eta}$, Client parameters $\tilde{\beta_1},\tilde{\beta}_2\in[0,1)$, Client adaptivity param. $\tilde{\epsilon}>0$, Client minibatch sampling rule and size\\
        \greenbg{Server parameter $\beta_1 \in [0,1)$}\\
        \bluebg{Server parameters $\beta_1,\beta_2\in[0,1)$,}\\
        \bluebg{Server adaptivity param. $\epsilon>0$}\\
        \ENSURE Trained shared weights $\pmb{\phi}^*$, personalized layer weights $\{\pmb{\psi}^{m*}\}_{m=1}^M$
        \STATE Initialize shared weights $\pmb{\phi}_0$
        \STATE On each client $m\in[M]$, initialize personalized layer weights $\pmb{\psi}^m_0$
        \STATE \greenbg{Initialize FedAvgMomentum state $\mb{m}_0 = \mb{0}$}
        \STATE \bluebg{Initialize FedAdam states $\mb{m}_0=\mb{0}, \mb{v}_0=\epsilon^2\mb{1}$}
        \FOR {server epochs $k = 1$ to $K$}
        \STATE Server sends $\pmb{\phi}_{k-1}$ to all clients
        \FOR {clients $m=1$ to $M$}
        \STATE Sample minibatch $\mcal{M}_m \subset \mcal{D}_m$
        \STATE Construct full weights $\tilde{\pmb{\theta}}_{0,k}^m = [\pmb{\phi}_{k-1},\pmb{\psi}^m_{k-1}]$
        \STATE Initialize Adam states $\mb{m}_{0,k}=\mb{0}, \mb{v}_{0,k}=\mb{0}$
        \\ \textit{Client Updates} :
        \FOR {client epochs $k'=1$ to $K'$}\label{line:PLClientStart}
        \STATE $\tilde{\mb{g}}_{k',k}^m = \nabla_{\pmb{\theta}} \frac{1}{|\mcal{M}_m|}\sum_{(\mb{X},Y)\in\mcal{M}_m}l(f_{\tilde{\pmb{\theta}}_{k'-1,k}}(\mb{X},Y))$
        \STATE $\mb{m}_{k',k}^m = \tilde{\beta}_1\mb{m}_{k'-1,k}^m + (1-\tilde{\beta}_1) \tilde{\mb{g}}_{k',k}^m$
        \STATE $\mb{v}_{k',k}^m = \tilde{\beta}_2\mb{v}_{k'-1,k}^m + (1-\tilde{\beta}_2)(\tilde{\mb{g}}_{k',k}^m)^{\odot 2} $
        \STATE $\mb{\hat{m}}_{k',k}^m = \mb{m}_{k',k}^m(1-\tilde{\beta}_1^{k'})^{-1}$
        \STATE $\mb{\hat{v}}_{k',k}^m = \mb{v}_{k',k}^m(1-\tilde{\beta}_2^{k'})^{-1}$
        \STATE $\tilde{\pmb{\theta}}_{k',k}^m = \tilde{\pmb{\theta}}_{k'-1,k}^m - \tilde{\eta}\mb{\hat{m}}_{k',k}^m\odot ((\mb{\hat{v}}_{k'-1,k}^m)^{\odot\frac{1}{2}}+\tilde{\epsilon}\mb{1})^{\odot -1}$
        \ENDFOR
        \STATE Extract $\pmb{\phi}^m_k$ and $\pmb{\psi}^m_k$ from $\tilde{\pmb{\theta}}^m_{K',k}$, i.e. $[\pmb{\phi}^m_k,\pmb{\psi}^m_k]=\tilde{\pmb{\theta}}^m_{K',k}$ and store $\pmb{\psi}^m_k$ locally
        \STATE Client sends $\tilde{\mb{g}}^m_k=\pmb{\phi}^m_{k-1}-{\pmb{\phi}}^m_{k}$ to server
        \ENDFOR\label{line:PLClientEnd}
        \\ \textit{Server Updates} :
        \STATE $\Delta_k = \sum_{m=1}^M \left( \frac{|\mcal{M}_m|}{\sum_{m'=1}^M |\mcal{M}_{m'}|} \right) \tilde{\mb{g}}^m_k $ \label{line:PLServerStart}
        \STATE \redbg{$\pmb{\phi}_k =   \pmb{\phi}_{k-1} - \eta\Delta_k$}
        \STATE \greenbg{$\mb{m}_k = \beta_1\mb{m}_{k-1} + (1-\beta_1)\Delta_k$}
        \STATE \greenbg{$\pmb{\phi}_k = \pmb{\phi}_{k-1} - \eta\mb{m}_k$}
        \STATE \bluebg{$\mb{m}_k = \beta_1\mb{m}_{k-1}+(1-\beta_1)\Delta_k$ }
        \STATE \bluebg{$\mb{v}_k = \beta_2\mb{v}_{k-1} + (1-\beta_2)\Delta_k^{\odot 2}$}
        \STATE \bluebg{$\pmb{\phi}_k = \pmb{\phi}_{k-1}-\eta \mb{m}_k \odot (\mb{v}_k^{\odot \frac{1}{2}}+\epsilon\mb{1})^{\odot -1}$}
        \ENDFOR
        \RETURN $\pmb{\phi}^*=\pmb{\phi}_K$, $\{ \pmb{\psi}^{m*} \}_{m=1}^M = \{ \pmb{\psi}_K^m \}_{m=1}^M$ \label{line:PLServerEnd} 
    \end{algorithmic} 
    \label{alg:pers-fl}
\end{algorithm}

\section{Exploratory Analysis of NREL ComStock Dataset}
\label{sec:exploratory}

\begin{table}[!tb]
    \centering
    \caption{Correlation of features with the energy consumption.}
    \begin{tabular}{|c|c|c|c|}
        \hline
        \textbf{Feature Name} & \textbf{California} & \textbf{Illinois} & \textbf{New York}\\
        \hline
        \multicolumn{4}{|c|}{\emph{Static features}}\\
        \hline
        Total floor space (sq. ft.) & 0.9591 &  0.9245 & 0.9245\\
        \hline
        \makecell{Cooling equipment\\capacity (tons)} & 0.9905 & 0.9683 & 0.9173 \\
        \hline
        \makecell{Heating equipment\\capacity ($kBTU/h$)} & -0.1853 & 0.9374 & 0.9374 \\
        \hline
        External wall area ($m^2$) & 0.3702 & 0.7385 & 0.7385\\
        \hline
        External window area ($m^2$) & 0.2417 & 0.9714 & 0.9714\\
        \hline
        Year built & -0.4462 & -0.2652 & -0.2652\\
        \hline
        \multicolumn{4}{|c|}{\emph{Time-varying features}}\\
        \hline
        Dry Bulb Temperature (${}^\circ C$) &0.5473 & 0.3789 & 0.2506 \\
        \hline
        \makecell{Global horizontal\\radiation ($W/m^2$)} & 0.4101 & 0.4870 & 0.4870 \\
        \hline
        \makecell{Direct normal\\radiation ($W/m^2$)} & 0.3838 & 0.3789 & 0.3789 \\
        \hline
        \makecell{Diffuse horizontal\\radiation ($W/m^2$)} & 0.3446 & 0.4577 & 0.4577 \\
        \hline
        Wind speed ($m/s$) & 0.2288 & 0.1396 & 0.1396 \\
        \hline
        Wind direction (deg) & -0.0246 & 0.0787 & 0.0787 \\
        \hline
        Relative humidity (\%) & -0.3853& -0.2794 & -0.2794 \\
        \hline
    \end{tabular}
    \label{tab:static}
\end{table}

For the purpose of experiments, we use the NREL ComStock dataset, which is available at \texttt{\url{https://data.openei.org/submissions/4520}}. It contains semi-synthetic energy consumption data for over 300,000 commercial buildings spread across each of the 50 US states and District of Columbia. In order to ensure that our results generalize, we choose 42 buildings each from three states: California, Illinois, and New York, which represent three geographically distinct datasets for all experiments. In the context of FL and PL-FL, we treat these 42 buildings as 42 clients, and experiments are carried out for each of the three-state datasets. The energy consumption data has a granularity of 15 minutes, and the available data has a time range spanning the year of 2018.

In order to choose features to be included in data points $\mb{x}_t$, we carry out an exploratory analysis of the dataset. It contains hundreds of features for each building, and therefore it is important to choose features which are predictive of the buildings' energy consumption. The available  features can be classified into two types: static and time-varying. Static features differ across buildings but are constant in time, and consist of building characteristics such as floor space, equipment rating, etc. On the other hand, time-varying features vary across both buildings and time, and include weather-related data such as temperature, wind speed, heat flux radiation, humidity, etc. We restrict the number of features to 8, out of which past energy consumption is one. Following the feature choices presented in~\cite{WK-etal:2019}, we choose the index of the 15-minute interval of the day (with values in $\{0,\cdots,95\}$) and that of the day of the week (with values in $\{0,\cdots,6\})$ as the next two features. Of the remaining 5 features, we assign 3 to be static and 2 to be time-varying. We randomly sample 20 buildings from each of the 3 datasets in order to measure correlation of available static and time-varying features with energy consumption. In order to prevent data contamination, the sampled buildings are not a part of the final 42 buildings used to train and test the STLF model.

\begin{figure*}[!tb]
    \centering
    \includegraphics[width=\linewidth]{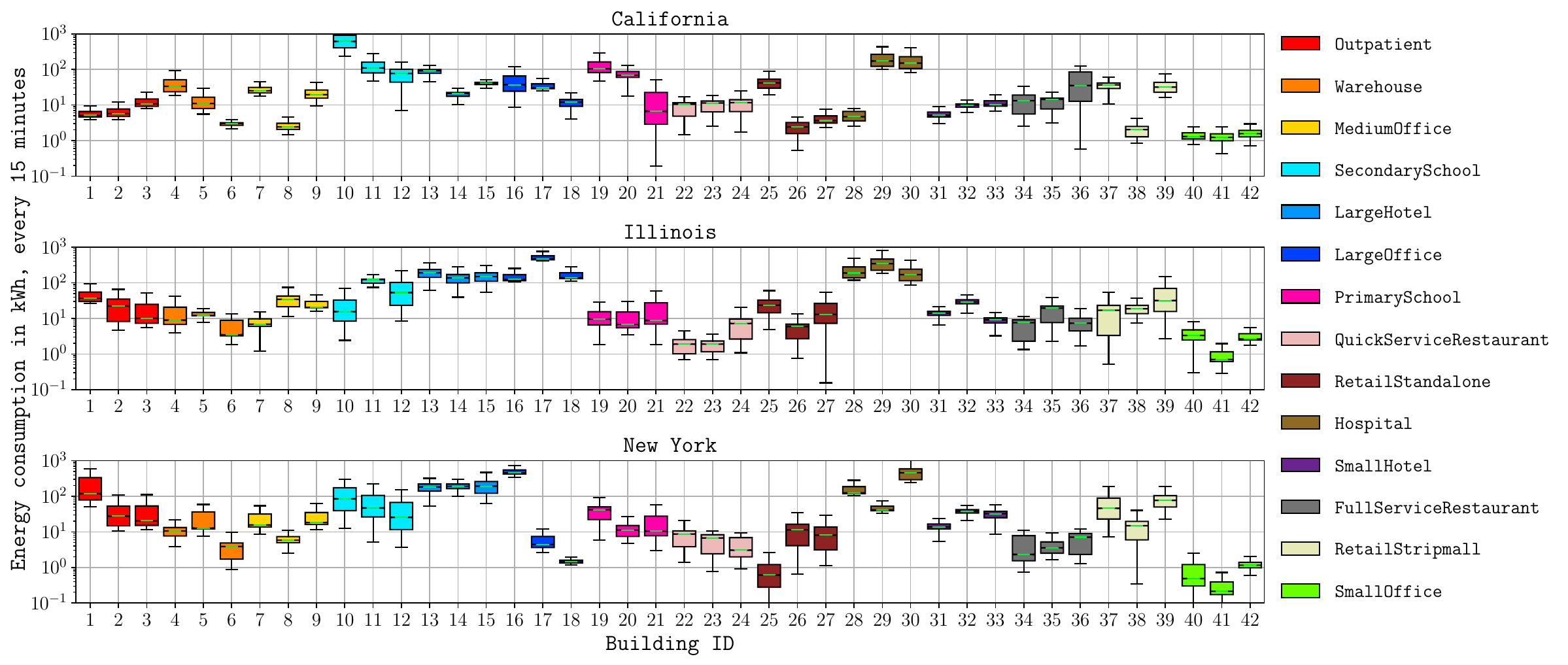}
    \caption{Box plots showing the heterogeneity (in terms of magnitude) of energy consumption for every building in all the three datasets.}
    \label{fig:hetero}
\end{figure*}

For the candidate static features, we average the energy consumption for each building with respect to time and calculate its Pearson correlation coefficient with said features. The resulting values demonstrate how well the static features predict the mean energy consumption of the sampled buildings, with the results presented in Table~\ref{tab:static}. From the table, it would be intuitive to choose total floor space and heating \& cooling equipment capacity as the static features. However, we consider the practicality of acquiring such data in a real-world setting. Utility companies in the US often possess architectural records of the buildings they provide electricity to, while they may not have updated data on the customers' heating and cooling equipment. Thus, we choose total floor space (sq.ft), external wall area ($m^2$), and external window area ($m^2$) as our static features. Interestingly, the California dataset does not follow the same trends as Illinois or New York, and this is evident from the correlations of heating equipment capacity and external wall area with energy consumption.

We carry out a similar analysis to choose the ideal time-varying features, with results shown in the lower half of Table~\ref{tab:static}. For each building, we first calculate the correlation of every feature with energy consumption, followed by averaging the correlation coefficients across the 20 buildings. From Table~\ref{tab:static}, an obvious choice would be to choose global horizontal radiation and direct normal radiation as the time-varying features, especially considering Illinois and New York. However, similar to static features, heat flux radiation is not a metric which is often measured by either the utility company or local meteorological organizations. Therefore they are excluded to maintain practicality of the STLF model. In their place, we include dry bulb temperature and wind speed as our time-varying features. Thus, $\mb{x}_t$ is laid out as
\begin{align*}
    \mb{x}_t = \bm{ \bt{energy consumption}_t\\
        \bt{time of day 0...95}_t\\
        \bt{day of week 0...6}_t\\
        \bt{dry bulb temperature}_t\\
        \bt{wind speed}_t\\
        \bt{total floor space}\\
        \bt{external wall area}\\
        \bt{external window area}
    }.
\end{align*}

\subsubsection*{Data Heterogeneity} While the correlation analysis of different features with energy consumption demonstrates the significant heterogeneity of data between different datasets, we are more interested in the internal heterogeneity among the 42 buildings of each dataset, since they act as the clients. To that end, a major indicator of heterogeneity is the magnitude of energy consumption of a given building. This is because ComStock contains 14 different types of buildings such as warehouses, schools, restaurants, etc., and our final dataset consists of 3 buildings randomly selected from each category. In order to visualize the heterogeneity, we create box plots showing variation of energy consumption in time for all 3 datasets in Figure~\ref{fig:hetero}. The sheer difference in energy consumption levels necessitates the $y$-axis to be logarithmic for proper visualization. While buildings of similar type have similar magnitudes of energy consumption, it varies significantly between buildings of different types. For example, large hotels in all the 3 datasets have greater energy consumption than small offices. 

Another form of heterogeneity manifests in the variance of energy consumption across time, which can be inferred by observing the spread of boxes in the plots. This can pose a significant challenge for STLF, since the underlying LSTM model will have to reliably learn the signals which indicate that energy consumption of a building is about to fall or rise significantly. From Figure~\ref{fig:hetero}, it can be observed that schools, restaurants, and retail have significant variance in energy consumption, which can be explained by the fact that a large amount of energy is consumed during operating hours whereas outside those hours the energy consumption becomes negligible. On the other hand, hospitals have smaller variance since they provide a large portion of their services around the clock. In the next section, we explore the effects of such heterogeneity on FL, and the remedial effects provided by PL-FL.

\section{Numerical Experiments}
\label{sec:experiments}

In this section, we carry out numerical experiments to demonstrate the effectiveness of personalization layers for addressing clients' data heterogeneity. Each experiment is carried out for all the 3 datasets. The energy consumption data for all clients is split into train, validation, and test sets along the time axis. The first $80\%$ of the data from each client constitutes the train set, and the next two chunks of $10\%$ each constitute the test and validation sets respectively. We normalize each feature of the train, validation, and test sets to the range $[0,1]$ by using min-max scaling, wherein the scaling factors are derived from the train set. In order to measure the accuracy of forecasts, we use two error metrics. Mean absolute error (MAE) is widely used for two time series $\{x_t\}_{t=1}^T$ and $\{\hat{x}_t\}_{t=1}^T$, which is given as
\begin{align*}
    \text{MAE} = \frac{1}{T}\sum_{t=1}^T |x_t - \hat{x}_t|.
\end{align*}
The second metric we use is called the mean absolute scaled error (MASE)~\cite{RJH-ABK:2006}, which is a scale-invariant error metric. MASE makes a further assumption that the time series $\{x_t\}_{t=1}^T$ is the ``ground truth'' while $\{\hat{x}_t\}_{t=1}^T$ is the output of a forecasting process. It is given as
\begin{align*}
    \text{MASE} = \frac{1}{T} \sum_{t=1}^T \frac{ |x_t-\hat{x}_t| }{\sum_{t'=2}^T |x_{t'} - x_{t'-1}|} = \frac{ \text{MAE} }{\sum_{t'=2}^T |x_{t'} - x_{t'-1}|}.
\end{align*}
MASE serves as an intuitive metric for evaluating the performance of a forecast model because if MASE is greater than one, then the given model in question is inferior to the na\"{i}ve technique of simply forecasting the last-known data point. We highlight that both MAE and MASE are calculated by rescaling the results to their normal scales, rather than on the $[0,1]$ scale.

\subsection{Model and Configurations} 

\begin{figure}[!tb] 
    \centering
    \subfloat[No personalization, i.e., FL.\label{fig:config1}]{%
        \includegraphics[width=0.5\linewidth]{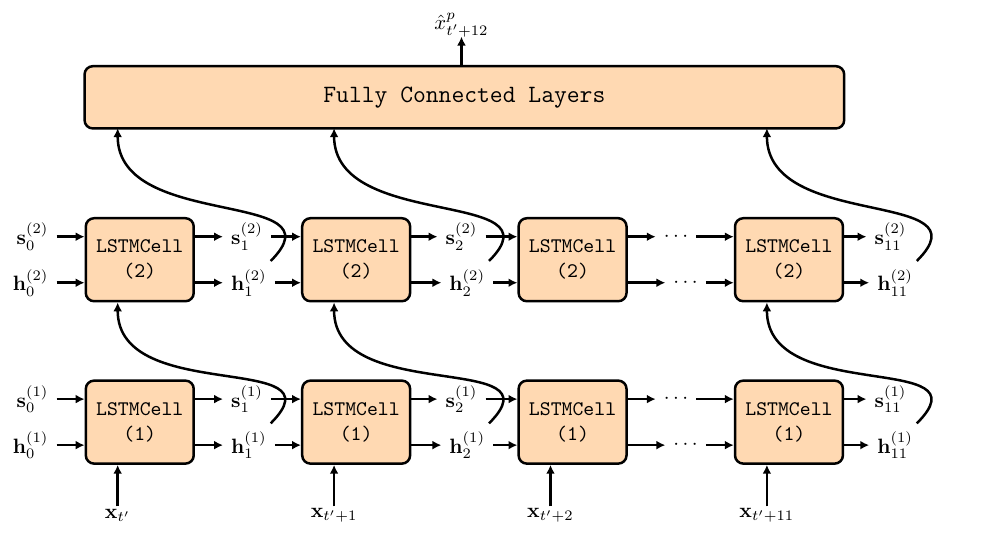}}
    \hfill
    \subfloat[PL-FL Configuration 1: The fully connected head is personalized.\label{fig:config2}]{%
        \includegraphics[width=0.5\linewidth]{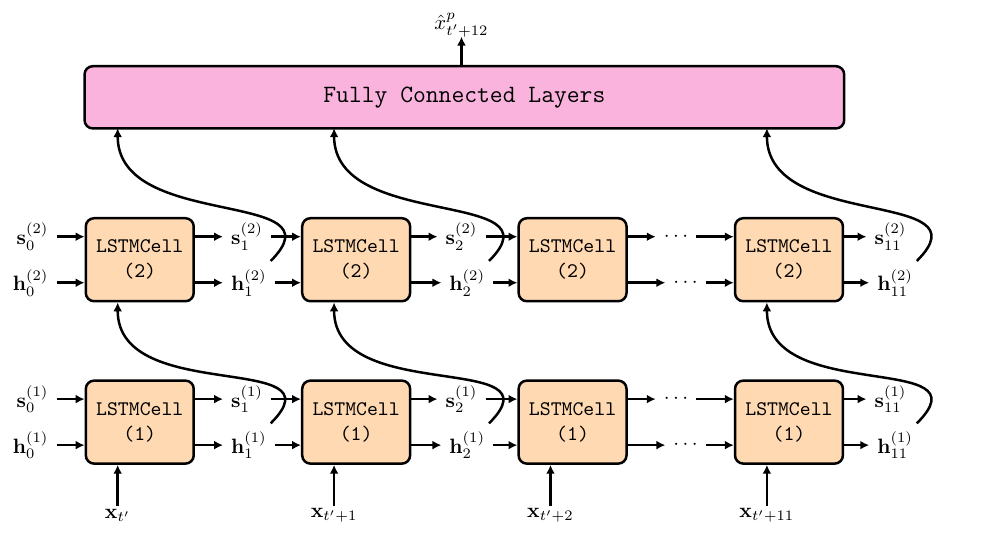}}
    \\
    \subfloat[PL-FL Configuration 2: The fully connected head and the top LSTM stack are personalized.\label{fig:config3}]{%
        \includegraphics[width=0.5\linewidth]{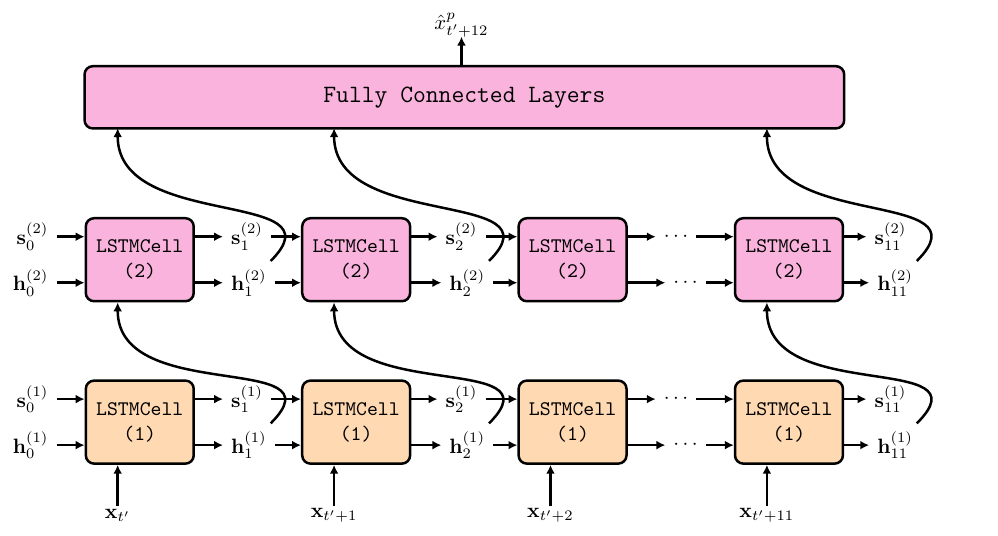}}
    \hfill
    \subfloat[PL-FL Configuration 3: The entire model is personalized.\label{fig:config4}]{%
        \includegraphics[width=0.5\linewidth]{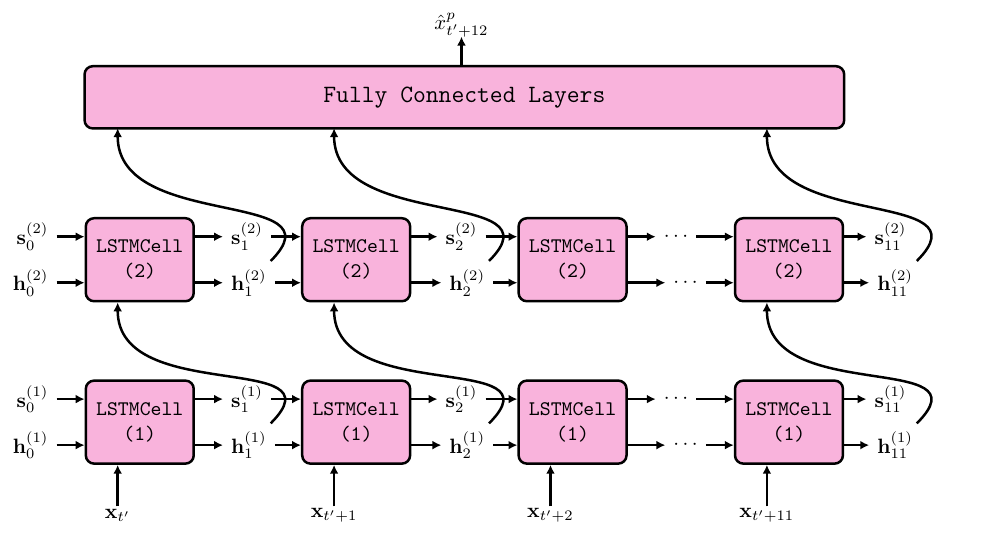}}
    \caption{Different personalization configurations for PL-FL. Layers colored \magbg{magenta} are personalization layers, while those colored \orbg{orange} are shared layers.}
    \label{fig:pers} 
\end{figure}
We use the two-stacked LSTM model shown in Figure~\ref{fig:lstm1}. After performing grid-search for optimal hyperparameters on the validation set, we choose the hyperparameters as reported in Table~\ref{tab:lstm-h}, which remain constant across all further experiments. For experiments involving PL-LF, we use three different configurations shown in Figure~\ref{fig:pers}. Configuration 1 represents the personalization of the fully connected head. Configuration 2 personalizes the top LSTM stack as well as the fully connected head. Configuration 3 represents complete personalization; that is,  each client trains locally and does not communicate with the server. In addition to FL and the three configurations of PL-FL, we consider the No-FL (denoting \emph{STLF with no FL}) model, wherein all the clients pool their data into a common dataset, which is then used to train the STLF model.
%

\subsection{Experiment Setup} 
\begin{table}[tb!]
    \centering
    \caption{Hyperparameters}
    \begin{tabular}{|c|c|}
        \hline
        \textbf{Parameters} & \textbf{Values}\\
        \hline
        Lookback ($T$) & 12\\
        \hline
        Stack 1 state size ($\text{dim}(\mb{h}^{(1)}), \text{dim}(\mb{s}^{(1)})$) & 20\\
        \hline
        Stack 2 state size ($\text{dim}(\mb{h}^{(2)}), \text{dim}(\mb{s}^{(2)})$) & 20\\
        \hline
        Fully Connected Layer Sizes & (240,120,60,1)\\
        \hline
        Fully Connected Layer Activation & PReLU \\
        \hline
        Batch Size & 64\\
        \hline
    \end{tabular}
    \label{tab:lstm-h}
\end{table}
All code is written in the APPFL package, which is a Python-based open-source federated learning framework~\cite{APPFL} developed at Argonne National Laboratory. In order to simulate a FL setup, APPFL uses the Message Passing Interface  protocol to initialize multiple parallel jobs, each of which simulates a federated client. All training and evaluations were carried out on the Swing and Bebop clusters at Argonne. The former consists of nodes with 8 Nvidia A100 GPUs, while the latter consists of nodes with dual Intel Xeon E5-2695v4 processors with 18 cores each.

\subsection{Choice of Server Algorithm}

\begin{table}[tb!]
    \setlength{\tabcolsep}{3pt}
    \centering
    \caption{Comparison between different server algorithms\\(averaged across 3 clients)}
    \begin{tabular}{|c|c|c|c|c|c|c|}
        \hline
        \multirow{2}{*}{\textbf{Dataset}} & \multicolumn{2}{c|}{\textbf{FedAvg}} & \multicolumn{2}{c|}{\textbf{FedAvgMomentum}} & \multicolumn{2}{c|}{\textbf{FedAdam}}\\
        \cline{2-7}
        & \textbf{MAE} & \textbf{MASE} & \textbf{MAE} & \textbf{MASE} & \textbf{MAE} & \textbf{MASE}\\
        \hline
        California & 0.3302 & 2.2622 & 0.3181 & 2.1246 & \textbf{0.2861} & \textbf{1.9540} \\
        \hline
        Illinois & \textbf{1.7534} & \textbf{2.6551} & 2.9235 & 4.3619 & 1.9070 & 2.8521 \\
        \hline
        New York & 9.8462 & 3.0918 & 10.4671 & 4.3502 & \textbf{8.3716} & \textbf{2.6585} \\
        \hline
    \end{tabular}
    \label{tab:server-algo}
    \setlength{\tabcolsep}{10pt}
\end{table}

\begin{figure*}[!tb]
    \centering
    \includegraphics[width=\linewidth]{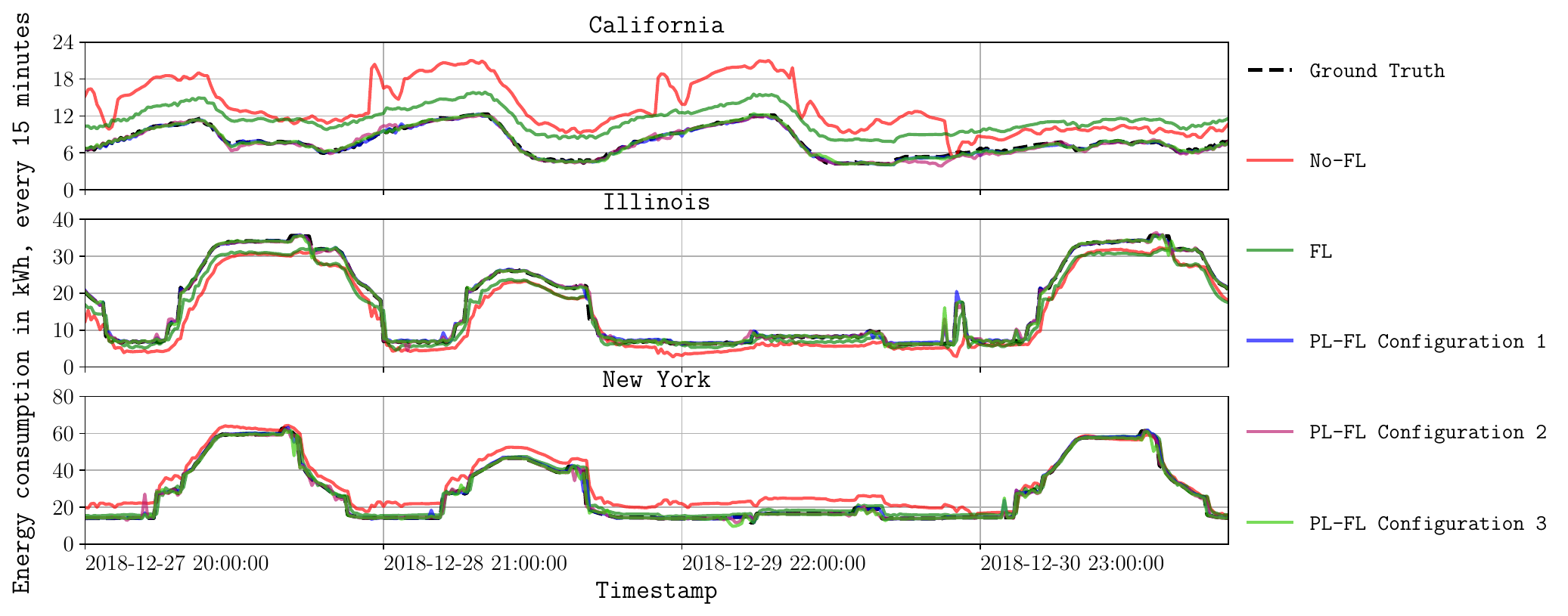}
    \caption{Forecasting performance of No-FL, FL, and PL-FL. The client shown here is client \#2 for each dataset. Shown here are the last 150 points from the test set.}
    \label{fig:forecast-fl}
\end{figure*}

Recall that both FL and PL-FL support three server algorithms: FedAvg, FedAvgMomentum, and FedAdam. In this experiment we compare the performance of each of these algorithms, and the one having the best performance is selected for carrying out comparisons between FL and PL-LF. We use the first 3 clients out of 42 from each dataset and train the STLF model with FL for a total of 2,000 server epochs (i.e., $K=2000$) and 4 client epochs for each server epoch (i.e., $K'=4$). For all the three algorithms, the client learning rate $\tilde{\eta}$ is fixed at 0.001, while on the server side, FedAvg and FedAvgMomentum use a learning rate of $\eta=1$ while FedAdam uses a learning rate of $\eta=0.01$. FedAdam uses $\beta_2=0.999$ while both FedAdam and FedAvgMomentum use $\beta_1=0.99$. Once training is completed for all the 9 models (3 datasets, with 3 algorithms per dataset), the resulting error metrics on the test set (averaged across the 3 clients) are reported in Table~\ref{tab:server-algo}. 

The results clearly indicate that FedAvg and FedAdam show superior performance to FedAvgMomentum across all 3 datasets. Within the former two, however, FedAdam provides better performance than does FedAvg on the California and New York datasets. These results concur with observations from non-federated settings, which note that since Adam and FedAdam are \emph{adaptive} algorithms~\cite{DPK-JB:2014} (i.e., they dynamically change per-parameter learning rates proportional to gradient values), they are robust to misspecifications of learning rate by the user. 
While FedAdam is more computationally expensive because of having to maintain the $\mb{m}$ and $\mb{v}$ state vectors, it is the superior choice compared with the alternatives.

We conclude by noting that while this experiment provides insight into the performance of server algorithms, the resulting models are not satisfactory from the perspective of STLF. The reason is that all MASE values in Table~\ref{tab:server-algo} are greater than 1. As discussed before, this implies that on an average, the na\"ive method of forecasting the last point will outperform all the models trained with FL. In the following experiment, we explore whether PL-FL can alleviate this issue.




\subsection{Comparison of Personalization Configurations} 

\begin{table*}[tb!]
    \centering
    \caption{Comparison of different STLF training algorithms (averaged across 42 clients)}
    \begin{tabular}{|c|c|c|c|c|c|c|c|c|c|c|}
        \hline
        \multirow{2}{*}{\textbf{Dataset}} & \multicolumn{2}{c|}{\textbf{No-FL}} & \multicolumn{2}{c|}{\textbf{FL}} &
        \multicolumn{2}{c|}{\makecell{\textbf{PL-FL}\\ \textbf{Configuration 1}}} & \multicolumn{2}{c|}{\makecell{\textbf{PL-FL}\\ \textbf{Configuration 2}}} & \multicolumn{2}{c|}{\makecell{\textbf{PL-FL}\\ \textbf{Configuration 3}}}\\
        \cline{2-11}
        & \textbf{MAE} & \textbf{MASE} & \textbf{MAE} & \textbf{MASE} & \textbf{MAE} & \textbf{MASE} & \textbf{MAE} & \textbf{MASE} & \textbf{MAE} & \textbf{MASE}\\
        \hline
        California & 21.3557 & 23.5128 & 22.5103 & 29.3381 & \textbf{20.7145} & \textbf{0.8443} & 20.7486 & 0.9201 & 20.7406 & 1.2528 \\
        \hline
        Illinois & 18.3858 & 5.3659 & 22.8306 & 22.3734 & \textbf{18.0419} & 0.7152 & 18.0878 & \textbf{0.6928} & 18.0989 & 0.6984 \\
        \hline
        New York & 25.7914 & 28.4246 & 26.7211 & 32.3417 & \textbf{24.8082} & 0.8449 & 24.8322 & 0.8340 & 24.8119 & \textbf{0.7571}\\
        \hline
    \end{tabular}
    \label{tab:methods}
\end{table*}

For this experiment we compare the performance of the three configurations of PL-FL shown in Figure~\ref{fig:pers} with FL and No-FL. We use the full dataset of 42 clients for all 3 datasets, which results in a total of 15 training runs. Both FL and the 3 configurations of PL-FL use the Adam optimizer at each client and the FedAdam algorithm at the server, with $K=2000$ server epochs with each having $K'=4$ client epochs. The learning rates and other parameters are same as the last experiment. For No-FL, which does not contain a federated server-client architecture and instead pools the data of all clients into a single dataset and trains on it, we set the number of epochs to be 8,000 for it to be commensurate with FL training algorithms. Further toward that end, we use the Adam optimizer for No-FL with a learning rate of $\eta=0.001$. 

The error metrics of the trained models on the clients' test set (averaged across all 42 clients) are reported in Table~\ref{tab:methods}.
From the table we see the superior performance of different PL-FL configurations over No-FL and FL. Specifically, configuration 1 of PL-FL, which involves personalizing the fully connected head, performs the best for all datasets with respect to the MAE metric. With respect to the MASE metric, each of the three configurations leads on one dataset each. 

The forecasting performance of all methods on a subset of points from the test set of client \#2 (same as last experiment) is shown in Figure~\ref{fig:forecast-fl}.
The qualitative benefit of PL-FL over No-FL and FL is also evident in the figure, wherein for all 3 datasets, FL and No-FL fail to forecast the correct magnitude of the true energy consumption. Since different clients contain energy consumption at significantly different magnitudes (see Figure~\ref{fig:hetero}), the clients with large or small magnitudes bias the STLF models trained with No-FL or FL, which is remedied with PF-FL. Furthermore, as opposed to the previous experiment, we see that except for PL-FL configuration 3 with California, every configuration of PL-FL achieves a MASE of less than 1 on all the three datasets. This implies that models trained with PL-FL have better forecasting capability than the na\"ive method, which is a requirement for forecasting models to be useful in practical applications. We also note that, overall, the present experiment has lower MASE errors but higher MAE errors than the previous experiment. This is because of the larger number of clients in the present experiment, with some clients having very high energy consumption such that small relative errors lead to a larger absolute error.

\subsection{Server-Client Communication Bandwidth}

\begin{table}[tb!]
    \centering
    \caption{Data exchange between server and clients per server epoch}
    \begin{tabular}{|c|c|c|}
        \hline
        \textbf{Algorithm} & \textbf{Parameters} & \textbf{Kilobits}\\
        \hline
        FL & 84362 & 2636 \\
        \hline
        PL-FL Config. 1 & 11520 & 360 \\
        \hline
        PL-FL Config. 2 & 4800 & 150 \\
        \hline
        PL-FL Config. 3 & 0 & 0\\
        \hline
    \end{tabular}
    \label{tab:data}
\end{table}

We conclude this section by discussing the amount of data transferred between server and clients during training. This topic can be contextualized by observing that personalization of different layers of a STLF model forms a spectrum. When none of the layers are personalized, we recover FL, while personalizing all the layers (i.e., configuration 3 of PL-FL) is equivalent to training a distinct forecasting model for each of the clients and can possibly be done locally. The utility company is therefore presented with a strategic choice in the amount of personalization it wants to implement. If the utility seeks to maximize forecast accuracy, then Table~\ref{tab:methods} suggests that in many cases the best strategy is partial personalization, rather than full federation or full personalization. On the other hand, there are other practical concerns such as data bandwidth available to smart meters. In Table~\ref{tab:data} we list the number of parameters (and the corresponding amount of data) that the server has to exchange with each client per server epoch. The number of parameters communicated is double the number of parameters in the shared layers, since the shared weights have to communicate twice (send and receive) per server epoch. The communication size was calculated considering each parameter to be a 32-bit float, as is the case with our experiments. For the present model, the fully connected head contains the largest number of parameters, and personalizing it results in the transferred data dropping from 2.363 MB to 360 KB. Personalizing the top LSTM stack results in a relatively modest drop from 360 KB to 150 KB. The exact trade-off between model accuracy and communication overhead will depend on a number of factors such as computational capabilities of the smart meter, available bandwidth, and required model accuracy and is ultimately an organization decision of the utility company depending on localized conditions.

\section{Conclusion}
\label{sec:conclusion}

In this paper we applied the concept of personalization layers to STLF in a federated setting. Using the well-studied LSTM model, we showed how different layers of this model can be personalized. We introduced PL-FL for the training of STLF models with personalization layers. Through experiments on the NREL ComStock dataset comprising clients with heterogeneous energy consumption data, we not only established the ideal choice of server algorithm for STLF in a federated setting but also established the superiority of PL-FL over its centralized and non-federated counterparts. A major takeaway from these  results is that data heterogeneity in FL can be remedied, and this observation serves as a step in the direction of practically deploying FL in the construction of STLF models. Future research will focus on the application of other privacy-preserving methods such as differential privacy in the personalization framework, as well as tailoring personalization to domains outside STLF.

\bibliographystyle{IEEEtran}
\bibliography{sources}
	
\end{document}